\documentclass{article}
\usepackage{ijcai25}

\usepackage{times}
\usepackage{soul}
\usepackage{url}
\usepackage[hidelinks]{hyperref}
\usepackage[utf8]{inputenc}
\usepackage[small]{caption}
\usepackage{graphicx}
\usepackage{amsthm}
\usepackage{booktabs}
\usepackage[switch]{lineno}
\usepackage{color}
\usepackage{amssymb} 
\usepackage{multirow}
\usepackage{subcaption}
\usepackage{pifont}
\usepackage{siunitx}
\usepackage{comment}
\usepackage{algorithm}
\usepackage{algpseudocode}
\usepackage{xcolor}
\usepackage{amsfonts, amsmath}
\usepackage{multirow}
\usepackage{tikz}
\usetikzlibrary{tikzmark}
\usepackage{graphicx}
\usepackage{calc}
\newcommand{\model}{DisUE}
\newcommand*\samethanks[1][\value{footnote}]{\footnotemark[#1]}

\title{Distilling A Universal Expert from Clustered Federated Learning}

\author{
Zeqi Leng$^{1,2}$
\and
Chunxu Zhang$^{1,2}$\thanks{Corresponding authors.}\and 
Guodong Long$^{3}$\and
Riting Xia$^4$\and
Bo Yang$^{1,2}$\samethanks
\\
\affiliations 
$^1$Key Laboratory of Symbolic Computation and Knowledge Engineering of Ministry of Education, China\\
$^2$College of Computer Science and Technology, Jilin University, China\\
$^3$Australian Artificial Intelligence Institute, FEIT, University of Technology Sydney\\
$^4$College of Computer Science, Inner Mongolia University, Hohhot, China\\
\emails
lengzq22@mails.jlu.edu.cn,
zhangchunxu@jlu.edu.cn, guodong.long@uts.edu.au\\
xiart@imu.edu.cn, ybo@jlu.edu.cn
}

\begin{document}

\maketitle

\begin{abstract}
Clustered Federated Learning (CFL) addresses the challenges posed by non-IID data by training multiple group- or cluster-specific expert models. However, existing methods often overlook the shared information across clusters, which represents the generalizable knowledge valuable to all participants in the Federated Learning (FL) system. To overcome this limitation, this paper introduces a novel FL framework that distills a universal expert model from the knowledge of multiple clusters. This universal expert captures globally shared information across all clients and is subsequently distributed to each client as the initialization for the next round of model training. The proposed FL framework operates in three iterative steps: (1) local model training at each client, (2) cluster-specific model aggregation, and (3) universal expert distillation. This three-step learning paradigm ensures the preservation of fine-grained non-IID characteristics while effectively incorporating shared knowledge across clusters. Compared to traditional gradient-based aggregation methods, the distillation-based model aggregation introduces greater flexibility in handling model heterogeneity and reduces conflicts among cluster-specific experts. Extensive experimental results demonstrate the superior performance of the proposed method across various scenarios, highlighting its potential to advance the state of CFL by balancing personalized and shared knowledge more effectively.
\end{abstract}

\section{Introduction}\label{intro}
Federated learning (FL) has emerged as a promising paradigm for privacy-preserving distributed training, enabling collaborative model development without exposing sensitive personal data~\cite{FL11,FL12,ijcaizhangcx}. The primary goal of FL is to train a high-quality consensus model through joint optimization. However, deploying FL across distributed clients is nontrivial due to the statistical heterogeneity of local data (i.e., non-IID). Standard averaging-based aggregation method~\cite{avg} tends to result in biased local updates, degrading the performance of the consensus model and slowing convergence. Prior works~\cite{scaffold,prox} have introduced correction mechanisms that align local and global update directions. While effective in reducing update bias, these approaches compromise local model personalization, a key factor in ensuring high performance under data heterogeneity.

Clustered Federated learning (CFL) effectively compensates for local personalization by training multiple cluster-level consensus models for heterogeneous devices~\cite{addcfl1,addcfl3,ma2023structured}. Unlike traditional FL,  CFL inherently preserves individual data characteristics through its clustering mechanism. Yet even state-of-the-art CFL approaches still face challenges in various practical settings. A key problem lies in the limited information flow across clusters: small clusters are prone to overfitting,  and large clusters typically converge to local optima. As a result, CFL consensus models tend to overfit local personalization, limiting their generalization capability.

In fact, neither traditional FL nor CFL methods effectively balance personalized knowledge with global consensus. An ideal solution aims to build a consensus model that preserves client-specific features. Yet, this is a significantly challenge, as naive aggregation of personalized knowledge tends to induce cluster-level model bias. These observations prompt us to explore a new question:

\textit{Can the consensus model be successfully conducted without undermining personalization?}

Transferring knowledge across heterogeneous clusters offers a potential solution. Recent advances in knowledge distillation (KD) have shown promise in facilitating such knowledge migration~\cite{kd}. However, a major limitation of traditional KD methods is their reliance on auxiliary datasets. The choice of auxiliary data significantly impacts model performance and introduces additional computational overhead~\cite{bridge}. These limitations motivate us to adopt a data-free knowledge distillation (DFKD) approach~\cite{fedgen}, in which a generator is trained through model inversion to synthesize pseudo-data on the fly, enabling knowledge transfer from the multiple teacher networks to the student network. 


To this end, we face several key challenges: \textbf{C1}. How to design a training method with multiple teacher networks such that heterogeneous knowledge remains non-conflicting? \textbf{C2}. How to effectively extract and ensemble cross-cluster knowledge, given that dynamic clustering induces random inter-cluster distribution shifts? \textbf{C3}. How to mitigate privacy risks in clustering, as the direct exposure of similarity values in existing CFL methods increases the risk of re-identification.

To simultaneously address all the aforementioned challenges, we propose a novel FL framework, {\model} (Distilled Universal Expert)), which enhances the consensus model while preserving  local personalization. {\model} follows a three-stage learning paradigm: local model training, iterative clustering, and universal expert distillation. In particular, to tackle \textbf{C1}, we construct a universal expert model from CFL and propose a category cluster-level knowledge migration mechanism to enable single expert-to-student distillation. To tackle \textbf{C2}, we design two adaptive components at the category cluster-level to handle dynamic inter-cluster heterogeneity. By leveraging category statistics, these components guide both model inversion and ensemble. To tackle \textbf{C3},  we introduce a lightweight similarity encryption protocol that prevents direct exposure of similarity values.

In a nutshell, our main contributions are as follows:
\begin{itemize}
\item \textbf{Novel Framework for FL}: We propose {\model}, a distillation-enhanced framework from CFL, which can maintain both intra- and inter-group knowledge in FL. 
\item \textbf{Adaptive Category Cluster-Distillation Mechanism}: We devise an adaptive Group Label Sampler (\textit{GLS}) and Group Weighting Factors (\textit{GWF}) to process heterogeneous knowledge across groups. It incorporates group-based distillation to reduce distillation aggregation overhead and globally shares a model to improve the performance of minority groups.
\item \textbf{Flexible plugin.} As a flexible algorithm that is orthogonal to existing CFL optimizers (IFCA, CFL, CFL-GP, PACFL). Our {\model} can easily enhance existing CFL methods, demonstrating the generality and compatibility of our inter-group aggregation mechanism.
\item \textbf{Superior performance.} Extensive experiments on three real-world datasets against diverse advancing baselines prove the consistently state-of-the-art performance of our proposed {\model}.
\end{itemize}

\begin{figure*}[htbp]
    \includegraphics[width=1.0\linewidth]{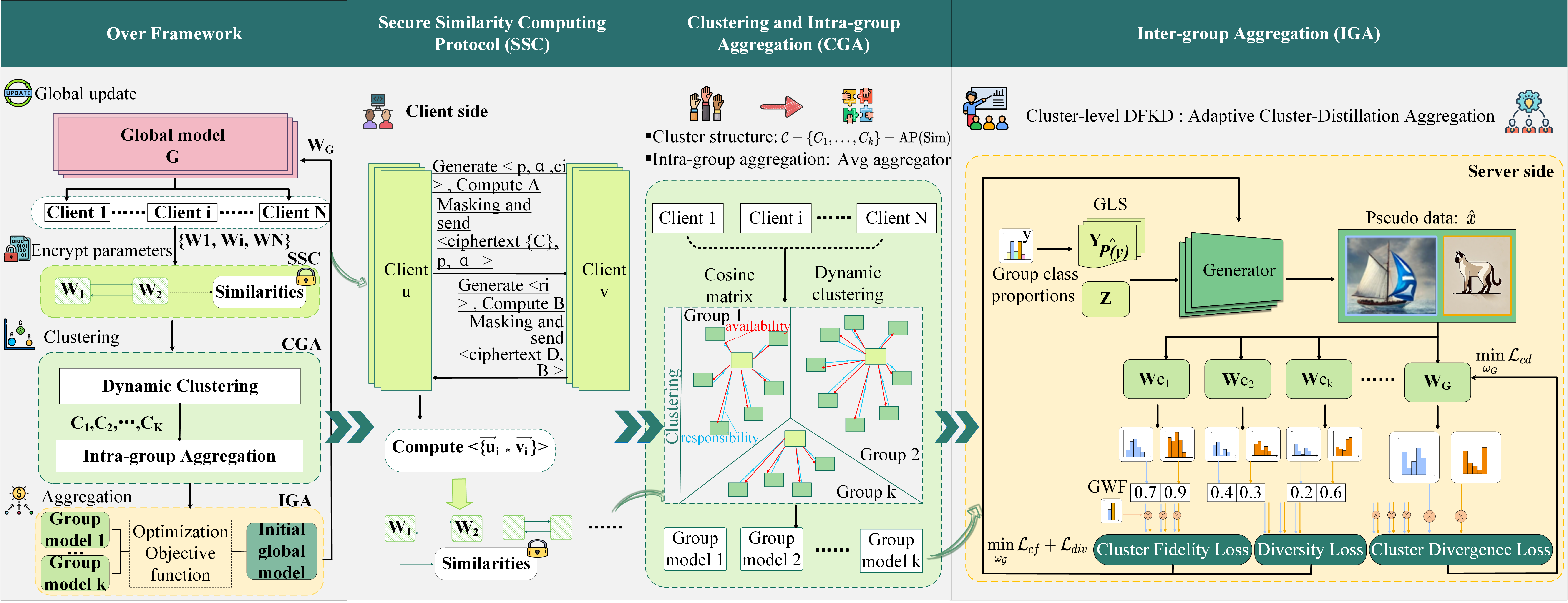}
    \caption{Overview of  {\model} workflow. (1) \textbf{Local Training}: Clients first train local models using private data. (2) \textbf{Secure Similarity Computing}: Clients employ the Secure Similarity Computing protocol to encrypt model parameters. (3) \textbf{Clustering and Intra-group Aggregation}: The server partition clients into clusters. Each cluster performs FedAvg averaging within its group. (4) \textbf{Inter-group Aggregation}: The framework distills a universal expert model while preserving data privacy.}
    \label{cfl}
\end{figure*}
\vspace{-5pt}
\section{Related Work}
\vspace{-3pt}
\subsection{Clustered Federated Learning}
CFL addresses statistical heterogeneity by grouping clients into clusters with homogeneous data distributions~\cite{long1,long2}. Existing CFL methods~\cite{ifca,cfl,addcfl2} primarily focus on client partitioning to derive accurate personalized expert models. Prior work has explored clustering under various assumptions, such as temporal domain shifts between training and test data~\cite{takingad}, joint label and feature skewness~\cite{drc}, and multiple types of data distribution shifts~\cite{fedsoft}. Other approaches optimize aggregation schedules~\cite{gp}. However, these methods restrict inter-cluster communication, causing expert models to overfit limited client data and suffer in generalizability. We bridge this gap by designing an expert model based on CFL, and improving its generalization via cluster-level DFKD transfer.
\vspace{-3pt}
\subsection{Data-free Knowledge Distillation}
\vspace{-3pt}
DFKD removes the need for auxiliary datasets in KD by transferring knowledge through model output exchange~\cite{yoo2019knowledge}. Current approaches fall into two categories based on source networks: single-teacher DFKD~\cite{shin2024teacher,feddf} and multi-teacher DFKD~\cite{hao2021data,ye2020data}. While most DFKD methods operate at the client level, they require more training rounds than aggregation methods~\cite{avg,scaffold} due to the larger scale of pseudo-data generated relative to original samples. To address this limitation, we introduce a cluster-level DFKD approach that lowers communication costs via group knowledge fusion.
\vspace{-5pt}
\section{Problem Statement}
\vspace{-3pt}
\textbf{CFL Pipeline.} CFL follows an iterative three-stage framework~\cite{liu2024casa}: 

\begin{itemize}
    \item \textit{Local training (L-phase)}. Each client performs local training and transmits model updates to the server. Let there be $N$ clients, where the $i$-th client holds a private dataset $D_i = \{(x_i^{(l)}, y_i^{(l)})\}_{l=1}^{n_i}$ with $n_i$ samples. Client $i$ updates its local parameters $\omega_i$ by minimizing the client-level empirical risk:
\begin{equation}
\min _{\omega_i} \frac{1}{n_i} \sum_{l=1}^{n_i} {f}\left(\omega_i;x_i^{(l)}, y_i^{(l)}\right)
\end{equation}
where \(f(\cdot)\) denotes the per-sample loss. All clients then send their parameters $\{\omega_i\}_{i=1}^N$ to the server.

\item \textit {Clustering (C-phase)}. The server partitions clients into $K$ disjoint clusters $\mathcal{C} = \{C_1,\ldots,C_K\}$ using similarity metrics (e.g., cosine distance).

\item \textit{Group model aggregation (G-phase)}. CFL maintains $K$ group models for $N$ clients. Within cluster $C_k$, $m$ clients with identically distributed data collaboratively train a group model $\omega_{C_k}$. We formulate the cluster-level optimization objective as:  
\begin{equation}\label{intra agg}
\min_{\omega_{C_k}}\frac{1}{m} \sum_{i=1}^{m} \mathcal{L}_{i}(\omega_{{i}};D_{i})
\end{equation}
Where $\mathcal{L}$ denotes the loss function for intra-group clients. The server subsequently distributes the updated cluster models $\{\omega_{C_k}\}_{k=1}^K$ to their respective clusters.
\end{itemize}

\noindent  \textbf{Emerging Requirements from CFL}. The core limitation arises from strict inter-cluster communication constraints. As prior analyses demonstrate, overcoming this challenge requires global information sharing – a critical need that drives our architectural redesign of the \textit{Group model aggregation}:

\begin{itemize}
    \item \textit{G-phase}: Directly applying existing methods to inter-cluster knowledge transfer encounters two fundamental barriers: (1) \textbf{Distribution Discrepancy}. The inherent heterogeneity across clusters in CFL leads to systematic bias when using traditional aggregation methods (e.g., FedAvg). (2) \textbf{Adaptation rigidity}. Static knowledge transfer mechanisms fail to respond to dynamic distribution shifts between clusters. These challenges demand a new CFL paradigm that supports adaptive transfer and dynamically adjusts to distributional changes induced by shifting cluster structures.
\end{itemize}
\vspace{-8pt}
\section{Methodology}
\vspace{-3pt}
In this section, we introduce {\model}, a cross-cluster federated learning framework that overcomes the limitations discussed in Section Introduction (Figure~\ref{cfl}). The framework operates through three core components: \textit{(1) \textit{Clustering and Intra-group Aggregation (CGA)}}: Partitions clients using cosine similarity-based clustering and extracts group-specific knowledge through intra-group model aggregation. \textit{(2) Inter-group Aggregation (IGA)}: Distills heterogeneous cluster expertise into a global model via data-free adversarial learning. \textit{(3) Secure Similarity Computing Protocol (SSC)}: Encrypts client parameters using PCSC, ensuring security against parameter leakage during similarity computation.
\vspace{-5pt}
\subsection{Clustering and Intra-group Aggregation}\label{clustering and intra-group agg}
\vspace{-3pt}
\textbf{Principles.} This stage achieves client dynamic clustering through angular similarity measurement, hyperparameter-free operation, and non-IID robustness. We quantify pairwise client relationships using cosine similarity between gradient updates. Unlike density-based methods~\cite{hdbscan,dbscan} that require predefined neighborhood thresholds, our design eliminates hyperparameters through message passing. Additionally, these algorithms exhibit unstable cluster structure under data heterogeneity. Our goal is to adopt a dynamic clustering mechanism that maintains robustness in extreme non-IID scenarios while enabling seamless integration with FL frameworks—critical for realizing a universal expert model. These requirements drive our adoption of Affinity Propagation (AP)~\cite{frey2007clustering}, which automatically identifies exemplar clients through iterative responsibility/availability updates.

\noindent  \textbf{Clustering.} We compute pairwise client similarities using cosine distance between parameter vectors $\omega_u$ and $\omega_v$:
\begin{equation}
\label{sim value}
\mathbf{Sim}(u, v)=\frac{\omega_u \cdot \omega_v}{\left\|\omega_u\right\|\left\|\omega_v\right\|}  
\end{equation}
where $\|\omega_u\|$ denotes the length of the parameter vector $\omega_u$. Let $\mathbf{Sim}$ be the cosine similarity matrix. The AP clustering obtains $K$ disjoint groups as follows:
\begin{equation}
\label{eq:clustering}
\mathcal{C} = \left\{C_1, \ldots, C_K\right\} = \text{AP}(\mathbf{Sim})
\end{equation}

\noindent  \textbf{Intra-group Aggregation}. For each cluster $C_j \in \mathcal{C}$, we aggregate member models through Eq.~\eqref{intra agg}, denotes as $\{\omega_{C_1}, \ldots, \omega_{C_K}\}$. Yielding specialist models $\{\omega_{C_j}\}_{j=1}^K$ that capture cluster-specific knowledge patterns.
\vspace{-5pt}
\subsection{Inter-group Aggregation}\label{sec:subsubDFKD}
\vspace{-3pt}
\textbf{Principles.} This stage addresses two co-existing challenges: adaptive cluster-wise label distribution shifts and cross-cluster distillation aggregation. Some aggregation methods based on subgroups are constrained by inter-group node symmetry or cluster topology symmetry~\cite{swiftagg,turbo}, while global averaging suppresses cluster-specific knowledge. Our solution breaks this stalemate through adaptive category group-level DFKD aggregation.  Specifically, (1) Dynamically adjusting cluster contributions through adaptive \emph{\textbf{GLS}} and \emph{\textbf{GWF}}, and (2) establishing cross-heterogeneous cluster knowledge transfer via prediction logit alignment between cluster experts and the global model. We fuse \emph{\textbf{GLS}} and \emph{\textbf{GWF}} into the cluster-level DFKD aggregation through a distillation-driven ensemble, creating a flexible mechanism for cross-cluster knowledge transfer. 

\noindent  \textbf{Handing Cluster-level Distribution Imbalance.}\label{Group Label Distribution Shift} Label distribution imbalance constitutes a primary manifestation of distributional discrepancies across heterogeneous clusters. The dynamic cluster assignment mechanism in CFL induces stochastic cluster-level label distribution variations across training rounds. Within cluster-level DFKD methods, this bias manifests through two critical pathways: (1) leveraging cluster-specific distribution characteristics to optimize generator training, and (2) aggregating cluster models according to their category-specific contributions. Inspired by ~\cite{ftg} and the underlying principle of counters, we proposed \emph{\textbf{GLS}} and \emph{\textbf{GWF}}.

The GLS samples representative labels guided by intra-cluster category statistics:
\begin{equation}
\hat{p}(y) \propto \sum_{k =1}^K \sum_{i=1}^{m} \mathbb{E}_{\left(x_i^{(l)}, y_i^{(l)}\right) \sim \mathcal{D}_i}\left[\mathbf{I}_{y_i=y}\right]=\sum_{k=1}^K n_{y}^{k}
\end{equation}
where $\mathbf{I}(\cdot)$ denotes the indicator function, and $n_{y}^{k}$ represents the cardinality of category $y$ samples in cluster $k$. The GW dynamically determines category importance weights per cluster:
\begin{equation}
\label{group weight factor}
\alpha^{k}_y = n_{y}^{k}/{\sum_{k=1}^K\sum_{i \in C_k}\; n_{y}^i}
\end{equation}
This adptaive weighting scheme ensures equitable integration of category-specific knowledge across clusters. These adaptive components regulate cluster-level knowledge transfer through accumulated category distribution statistics, enabling effective handling of non-IID data scenarios.

\noindent  \textbf{Cluster-level DFKD Aggregation Mechanism.} Our framework aims to facilitate cross-cluster knowledge transfer through data-free distillation, enabling the global model $\omega_G$ to assimilate heterogeneous cluster characteristics. We preserve inter-group knowledge foundations by maintaining cluster-averaged feature representations through aggregated cluster models serving as student networks. These cluster models subsequently function as teacher networks to supervise student model distillation.

The teacher-student knowledge transfer occurs through data-free distillation where teacher models guide student models to capture their specific data distributions. However, client-side generator deployment imposes prohibitive computational overhead on edge devices. To overcome this limitation, we design a generator  $\mathcal{G}$ at the server that synthesizes pseudo-samples $\hat{x}$ conditioned on category labels $y$:
\begin{equation}
\label{genERATOR}
\hat{x}=\mathcal{G}(\theta_{\mathcal{G}};z,y)
\end{equation}
where $z \sim \mathcal{N}(\mathbf{0},\,\mathbf{1})$ represents random Gaussian noise vectors, and $\theta_{\mathcal{G}}$ denotes the generator's parameters.

The cross-cluster distillation method achieves knowledge transfer through prediction discrepancy minimization between ensemble teachers and student network on generated samples, effectively capturing teacher-specific distribution patterns. We formulate the cluster distillation objective as:
\begin{equation}
\label{loss nd}
\min _{\omega_G} {\mathcal{L}_{cd}}:=  \mathbb{E}_{y \sim \hat{p}(y)} \mathbb{E}_{z \sim \mathcal{N}(\mathbf{0}, \mathbf{1})} \left[ \sum_{k=1}^K \alpha^{k}_y {\mathcal{L}^k_{c d}}\right]
\end{equation}
where $\hat{p}(y)$ denotes the cluster-level label distribution from Eq.~(5), and $\alpha^{k}_y$ represents the adaptive cluster weights defined in Eq.~(6) . To enable precise cluster knowledge transfer, we leverage the global model baseline while applying cluster-specific regulation through:
\begin{equation}
\label{loss k nd}
\mathcal{L}_{cd}^{k}= KL\left(\sigma\left( \mathbb{C}\left(\omega_{C_k} ;\hat{x} \right)\right) \| \sigma\left(\mathbb{C}\left( \omega_G; \hat{x}\right)\right)\right)
\end{equation}
Here $\mathbb{C}(\cdot)$ denotes the classifier output layer, $\sigma(\cdot)$ the softmax function, and $KL$ the Kullback-Leibler divergence measuring prediction distribution discrepancies.

To maintain semantic coherence in cross-cluster distillation while mitigating computational overhead, we optimize the centralized generator through cluster-conditioned cross-entropy minimization. The objective function for cluster fidelity, denoted as ${\mathcal{L}_{cf}}$, is given by:

\begin{equation}
\label{loss fid}
\min _{\theta_{\mathcal{G}}} {\mathcal{L}_{cf}}:=  \mathbb{E}_{y \sim \hat{p}(y)} \mathbb{E}_{z \sim \mathcal{N}(\mathbf{0}, \mathbf{1})} \left[ \sum_{k=1}^K \alpha^{k}_y {\mathcal{L}^k_{cf}}\right]
\end{equation}

The per-cluster fidelity loss is computed as:
\begin{equation}
\mathcal{L}_{cf}^k=\sum_{{k}=1}^K \alpha^{k}_y CE\left(\sigma\left(\mathbb{C}\left(\widetilde{x} ; \omega_k\right)\right), y\right)
\end{equation}

In CFL environments, cluster-specific knowledge exhibits multi-category characteristics. We adopt the diversity-aware regularization term $\mathcal{L}_{div}$~\cite{2021data} to enhance knowledge transfer through maximizing inter-sample dissimilarity.
\begin{equation}
\label{div}
\mathcal{L}_{d i v}=e^{\frac{1}{Q * Q} \sum_{i, j \in\{1, \ldots, Q\}}\left(-\left\|\hat{x}_i-\hat{x}_j\right\|_2 *\left\|z_i-z_j\right\|_2\right)}
\end{equation}
where $Q$ is the number of pseudo-samples, and $z_i$ denotes the noise of the $i$-th pseudo-sample.
\vspace{-3pt}
\subsection{Secure Similarity Computation}
\vspace{-3pt}
\noindent \textbf{Principles.} Similarity matrix exposure introduces re-identification risks, as adversaries could infer sensitive client attributes through parametric similarity analysis. We address this challenge through a lightweight \emph{Privacy-Preserving Cosine Similarity Computing (PCSC) protocol} protocol for CFL~\cite{2014toward}. The protocol prevents server access to raw similarity values through encrypted operations.
\vspace{-3pt}
\subsection{Optimization and Algorithm of {\model}}\label{op and algo}
\vspace{-3pt}
\textbf{Optimization Objective.}  Under CFL assumptions, intra-cluster data homogeneity allows cluster-optimal solutions via FedAvg, whereas inter-group label shifts degrade expert model performance. We seek a globally optimal solution $\omega_G$ that harmonizes $k$ cluster-specific distributions.

The {\model} framework establishes an adversarial game between the global model $\omega_G$ and a generator $\mathcal{G}$ (parameterized by $\theta_{\mathcal{G}}$), governed by a composite loss ($\mathcal{L}_{cd}$, $\mathcal{L}_{cf}$ and $\mathcal{L}_{div}$) that coordinates cross-cluster knowledge distillation through alternating optimization phases: (1) \textit{Maximization Phase} where $\mathcal{G}$ synthesizes hard-classifiable samples from cluster to amplify prediction conflicts, and (2) \textit{Minimization Phase} where $\omega_G$ aligns predictions with cluster models using boundary samples. This process progressively sharpens global decision boundaries while absorbing cluster-specific knowledge.

The unified optimization objective combines three loss components from Eqs.~\eqref{loss nd}, \eqref{loss fid}, and \eqref{div}:
\begin{equation}
\label{main function}
\min_{\omega_G}\;\max_{\theta_\mathcal{G}}\;\;
\mathbb{E}_{y \sim \hat{p}(y),\,z \sim \mathcal{N}(\mathbf{0}, \mathbf{1})}
\bigl(
\mathcal{L}_{cd}
\;+\;\beta_{cf}\,\mathcal{L}_{cf}
\;+\;\beta_{div}\,\mathcal{L}_{div}
\bigr)
\end{equation}
where $\beta_{cf}$ and $\beta_{div}$ balance the loss terms. This minimax optimization maximizes cross-cluster knowledge transfer, ultimately producing a universal model with consistent performance across clusters.
\begin{algorithm}[!t]
\caption{Workflow of {\model}} 
\label{algworkflow}
\begin{algorithmic}[1]
\Require 
    Initial global model $\omega_G^{(0)}$, 
    client set $\mathcal{N}_s$, 
    rounds $T$, 
    security parameter=$\mathsf{sec\_params}$
\Ensure 
    Final global model $\omega_G^{(T)}$, 
    generator parameters $\theta_{\mathcal{G}}^{(T)}$
    
\Statex \textbf{Initialization:}
\State Broadcast $\omega_G^{(0)}$ to all clients in $\mathcal{N}_s$

\For{round $t = 1$ \textbf{to} $T$}
    \Statex \textcolor{gray}{\textit{L-Phase: Local Training}}
    \State \textbf{Client Update}
    \ForAll{client $i \in \mathcal{N}_s$ \textbf{in parallel}}
        \State Local training: $\omega_i^{(t)} \leftarrow (\omega_G^{(t-1)})$
        \State $\mathsf{E}(\omega_i^{(t)}) \leftarrow \mathsf{SSC.Encrypt}(\mathsf{sec\_params}, \omega_i^{(t)})$
        \State Upload $\mathsf{E}(\omega_i^{(t)})$ to server
    \EndFor
    \State Initialize similarity matrix $\mathbf{Sim}^{(t)} \leftarrow \mathbf{0}_{|\mathcal{N}_s| \times |\mathcal{N}_s|}$
    \ForAll{pairs $(u, v) \in \mathcal{N}_s \times \mathcal{N}_s$ where $u < v$}
        \State Compute encrypted similarity: 
        \State $\quad Sim_{u,v}^{(t)} \leftarrow \mathsf{SSC.Compute}\big(\mathsf{E}(\omega_u^{(t)}), \mathsf{E}(\omega_v^{(t)})\big)$
        \State Update $\mathbf{Sim}^{(t)}[u,v] \leftarrow Sim_{u,v}^{(t)}$
    \EndFor
    
    \Statex \textcolor{gray}{\textit{C-Phase: Clustering}}
    \State $\{\omega_1^{(t)}, \ldots, \omega_K^{(t)}\} \leftarrow \mathsf{CGA\text{-}Clustering}(\mathbf{Sim}^{(t)})$
    \State Intra-group aggregation: 
    \State $\{\omega_{C_1^{(t)}}, \ldots, \omega_{C_K^{(t)}}\} \leftarrow \mathsf{CGA\text{-}Intra}(\{C_1^{(t)}, \ldots, C_K^{(t)}\})$
    
    \Statex \textcolor{gray}{\textit{G-Phase: Group Model Aggregation}}
    \State Global aggregation:
    \State $\omega_G^{(t)} \leftarrow \mathsf{FedAvg}(\omega_{C_1^{(t)}}, \ldots, \omega_{C_K^{(t)}})$
    \State Inter-group aggregation: 
    \State $\quad \omega_G^{(t+1)} \leftarrow \mathsf{IGA}\big(\{\omega_{C_1^{(t)}}, \ldots, \omega_{C_K^{(t)}}\}, \omega_G^{(t)}, \theta_{\mathcal{G}}\big)$
    \State $\mathsf{Server Update}(\omega_G^{(t+1)}, \theta_{\mathcal{G}})$
\EndFor

\State \Return $\omega_G^{(T)}, \theta_{\mathcal{G}}^{(T)}$
\end{algorithmic}
\end{algorithm}

\begin{table*}[!t]
    \centering
    \resizebox{0.9\textwidth}{!}{%
    \begin{tabular}{@{} ll *{3}{S[table-format=2.2]} *{3}{S[table-format=2.2]} *{3}{S[table-format=2.2]} @{}}
        \toprule[1.5pt]
         \multirow{2}{*}{\textbf{Type}} &\multirow{2}{*}{\textbf{Method}} 
            & \multicolumn{3}{c}{\textbf{SVHN}} 
            & \multicolumn{3}{c}{\textbf{CIFAR10}} 
            & \multicolumn{3}{c@{}}{\textbf{CIFAR100}} \\
        \cmidrule(lr){3-5} \cmidrule(lr){6-8} \cmidrule(l){9-11}
            & & \multicolumn{1}{c}{IID} & \multicolumn{1}{c}{0.1} & \multicolumn{1}{c}{0.01} 
            & \multicolumn{1}{c}{IID} & \multicolumn{1}{c}{0.1} & \multicolumn{1}{c}{0.01} 
            & \multicolumn{1}{c}{IID} & \multicolumn{1}{c}{0.1} & \multicolumn{1}{c@{}}{0.01} \\
        \midrule[1pt]
        
        \multirow{4}{*}{CFL} 
          & IFCA    & 91.66 & 88.63 & \underline{81.68} & 78.06 & 66.25 & 53.86 & 41.09 & 42.34 & 30.86 \\
          & CFL     & 80.05 & 69.03 & 68.16 & 69.29 & 55.37 & 47.60 & 39.65 & 38.28 & 26.33 \\
          & CFL-GP  & 79.95 & 71.53 & 69.01 & 71.09 & 55.30 & 48.56 & 41.22 & 34.13 & 26.75 \\
          & PACFL   & 93.09 & 87.46 & 80.65 & 78.30 & 67.26 & \underline{55.89} & 34.37 & \underline{43.14} & \underline{30.96} \\

        \addlinespace[3pt]
        \midrule[0.8pt]
        
        \multirow{2}{*}{DFKD}
          & FedDF   & 94.91 & 89.80 & 60.85 & 81.02 & 71.81 & 45.79 & 44.53 & 39.79 & 7.08 \\
          & FedFTG  & 94.73 & 90.89 & 74.92 & 81.67 & 73.19 & 51.61 & 42.71 & 36.61 & 30.85 \\
        
        \addlinespace[3pt]
        \midrule[0.8pt]
        
        \multirow{5}{*}{FL}
          & FedAvg  & 94.80 & 90.86 & 73.83 & 81.28 & 74.49 & 51.90 & 44.43 & 42.08 & 30.92 \\
          & FedProx & 94.93 & 90.52 & 73.61 & 81.77 & 74.96 & 51.86 & 44.27 & 41.42 & 30.06 \\
          & FedDyn  & \underline{95.10} & 90.59 & 78.95 & 81.95 & 75.01 & 51.50 & 45.44 & 40.64 & 16.13 \\
          & MOON    & 95.06 & \underline{91.62} & 76.94 & 81.23 & 73.96 & 51.45 & 42.65 & 41.23 & 30.07 \\
          & SCAFFOLD& 95.08 & 91.18 & 69.46 & \underline{82.02} & \underline{75.39} & 48.85 & \underline{49.55} & 25.32 & 17.84 \\
       
        \midrule[1.2pt]
        \multicolumn{2}{@{}l}{\textbf{DisUE}} 
            & \textbf{95.99} & \textbf{93.25} & \textbf{83.04} 
            & \textbf{83.25} & \textbf{76.22} & \textbf{58.83} 
            & \textbf{51.80} & \textbf{44.41} & \textbf{31.87} \\
        \bottomrule[1.5pt]
    \end{tabular}}
    \caption{Comparative Performance Analysis of Federated Learning Methods. Best results in \textbf{bold}, second best \underline{underlined}.}
    \label{experiments}
\end{table*}
\noindent \textbf{Algorithm.} 
Algorithm~\ref{algworkflow} illustrates the workflow of {\model}. We begin by broadcasting an initial global model $\omega_G^{(0)}$ to all clients. In each communication round $t$, clients first perform \textit{L-phase} on their private data. Subsequently, each client encrypts its updated parameters using the \textbf{SSC} protocol and transmits these encrypted updates to the server. The server obtains a similarity matrix $\mathbf{Sim}^{(t)}$, which it leverages to execute \textbf{CGA-Clustering}, partitioning clients into $K$ groups. Within group, parameters are aggregated via \textbf{CGA-Intra}, conducing clustering structure $\{\omega_{C_1^t}, \ldots, \omega_{C_K^t}\}$. 

Next, the global model $\omega_G^t$ is obtained via a federated aggregation (\textbf{FedAvg}) of these group models. To enable effective cross-cluster distillation, the \textbf{IGA} aggregates distilled knowledge from group models by refining $\omega_G^t$ into $\omega_G^{t+1}$ without direct access to raw client data. Finally, the server updates the global parameters (and the generator parameters $\theta_{\mathcal{G}}$ before proceeding to the next communication round. After $T$ rounds, {\model} returns the $\omega_G^{(T)}$ and $\theta_{\mathcal{G}}^{(T)}$.

\vspace{-8pt}
\begin{figure*}[!t]
\centering
\subfloat[Test accuracy on dir$(\epsilon=0.01)$]{
    \label{dir}
    \includegraphics[width=0.3\linewidth]{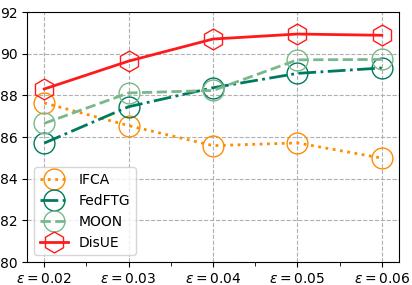}}
\hfill 
\subfloat[Test accuracy on act$(\cdot)$, dir$\epsilon=0.01$]{ 
    \label{act}
    \includegraphics[width=0.3\linewidth]{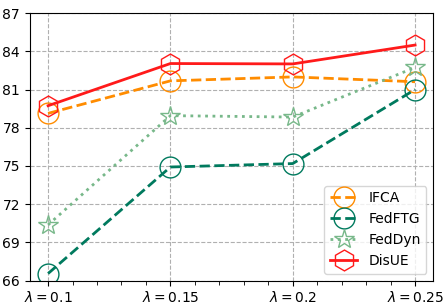}}
\hfill
\subfloat[Test accuracy on SVHN dir$(\epsilon=0.01)$]{ 
    \label{comm}
    \includegraphics[width=0.3\linewidth]{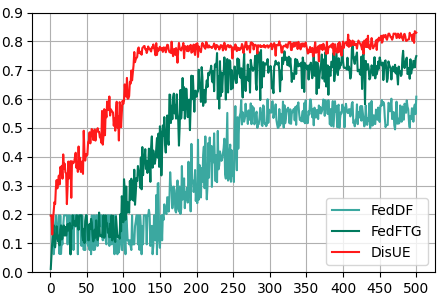}}
\caption{ (a) Test accuracy versus data heterogeneity measured by $\epsilon(\cdot)$. (b) Test accuracy versus the fraction $C$ of active clients per round ($\epsilon = 0.01$). (c) Test accuracy over 500 communication rounds (learning curve) with $\epsilon = 0.01$. All experiments are conducted on SVHN.}
\label{fig_6}
\end{figure*}
\vspace{-5pt}
\section{Experiments}
\vspace{-3pt}
\subsection{Implementation Details}\label{details}
\vspace{-3pt}
\textbf{Datasets.} We evaluate {\model} on three standard benchmarks: SVHN~\cite{svhn}, CIFAR-10~\cite{cifar}, and CIFAR-100~\cite{cifar}. To simulate federated data heterogeneity, we partition datasets using the Dirichlet distribution $Dir(\epsilon)$~\cite{hsu2019measuring}, with concentration parameters $\epsilon \in [0.01, 0.06]$ controlling non-IID difficulty levels. Lower $\epsilon$ values create more skewed client data distributions, representing challenging FL scenarios.

\noindent \textbf{Baselines.} We benchmark {\model} against 11 state-of-the-art FL methods spanning three key categories: (1) CFL methods including IFCA~\cite{ifca}, CFL~\cite{cfl}, CFL-GP~\cite{gp}, and PACFL~\cite{addcfl4}; (2) DFKD approaches FedDF~\cite{feddf} and FedFTG~\cite{ftg}; and (3) conventional FL methods covering FedAvg~\cite{avg}, FedProx~\cite{prox}, FedDyn~\cite{feddyn}, MOON~\cite{moon}, and SCAFFOLD~\cite{scaffold}. 

\noindent \textbf{Configurations.} For all methods, we set the communication rounds \( T = 500 \), the number of clients \( N = 100 \), with an active fraction \( Act = 0.15 \). For local training, we set the number of local epochs \( \text{localE} = 5 \), batch size = 50, and the weight decay to \( 1 \times 10^{-3} \). The learning rates for the classifier and generator are initialized to \( 0.1 \) and \( 0.01 \), respectively. The dimension \( z \) is set to 100 for CIFAR-10 and SVHN, and 256 for CIFAR-100. Unless otherwise specified, we adopt \( \beta_{\text{cf}} = 1.0 \) and \( \beta_{\text{div}} = 1.0 \). All our experimental results represent the average over five random seeds. For the classifier, we adopt the network architecture from ~\cite{he_Zhang_Ren_Sun_2016}. The generator architecture from \cite{ftg} is employed for both FedFTG and FedDF. 
\vspace{-5pt}
\subsection{Performance Comparison}\label{sota}
\vspace{-3pt}
\textbf{Test Accuracy}. Table ~\ref{experiments} reports the test accuracy achieved by all methods on the SVHN, CIFAR10, and CIFAR100 datasets. \textit {\textbf{(1)}} Among all scenarios, our method consistently outperforms the three baseline algorithm types on both IID and non-IID settings. \textit {\textbf{(2)}} Compared to CFL algorithms, {\model} achieves significantly better performance, primarily due to cross-group knowledge transfer, which enables its global model to leverage more comprehensive information. \textit {\textbf{(3)}} Compared to the DFKD algorithms, performance is further improved. This is attributed to the group-based distillation protocol having a pre-trained component, clustering, thereby mitigating the non-IID problem in advance. \textit {\textbf{(4)}} Experimental observations reveal distinct performance patterns across algorithms under varying data distributions. Global FL algorithm achieves superior accuracy in IID and $Dir(\epsilon=0.1)$. This phenomenon arises from the relatively balanced category distributions across clients, which induces aligned parameter update directions that undermine clustering efficacy. Conversely, under $Dir(\epsilon=0.01)$ non-IID conditions,  CFL demonstrates better performance over global FL.

\noindent \textbf{Communication Overhead}. We assessed the communication rounds needed for convergence by DFKD SOTA methods on SVHN with on-IID degrees set to \textit{0.01} in Figure.~\ref{comm}. \textit {\textbf{(1)}} Our method achieves accelerated convergence rates through cluster-level adversarial distillation. \textit {\textbf{(2)}} During the first 100 iterations, our methods demonstrate greater performance gains compared to other DFKD approaches, evidencing that integration clustering effectively mitigates data heterogeneity issues and prevents early-stage local noise.

\noindent \textbf{Data Heterogeneity}. We evaluate the robustness of algorithms under varying degrees of non-IID data by measuring the test accuracy of all methods with non-IID degrees set to (0.02, 0.03, 0.04, 0.05, 0.06). Figure~\ref{dir} illustrates the test accuracy as a function of varying $Dir(\cdot)$, respectively showing the best-performing algorithm for each type of method. \textit{\textbf{(1)}}{\model} outperforms baseline methods across diverse configurations, demonstrating the effectiveness of CFL's clustering mechanism in (i) preserving heterogeneous data characteristics through clustering and (ii) maximizing knowledge utility via inter-group distillation. \textit{\textbf{(2)}} Global FL algorithms exhibit performance improvements with increasing Dirichlet coefficient $\epsilon$. \textit{\textbf{(3)}} CFL performance degrades progressively with higher values due to accumulating clustering errors.

\begin{figure*}[!t]
\centering
\subfloat[$\mathcal{L}_{cf}$]{
    \label{fig:a}
    \includegraphics[width=0.32\linewidth]{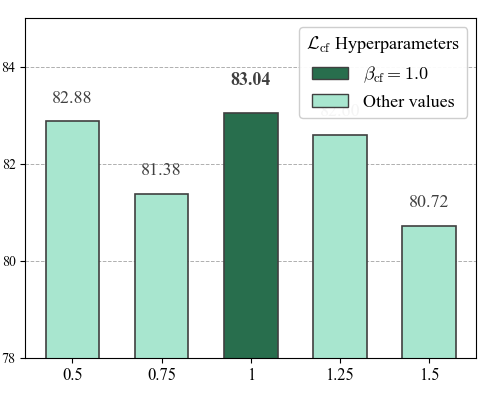}}
\subfloat[$\mathcal{L}_{div}$]{
    \label{fig:b}
    \includegraphics[width=0.32\linewidth]{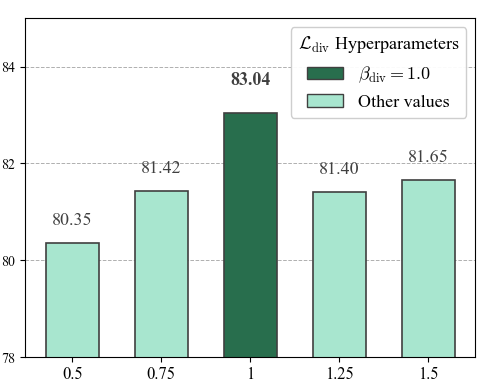}}
\subfloat[noise dimension]{
    \label{nz}
    \includegraphics[width=0.32\linewidth]{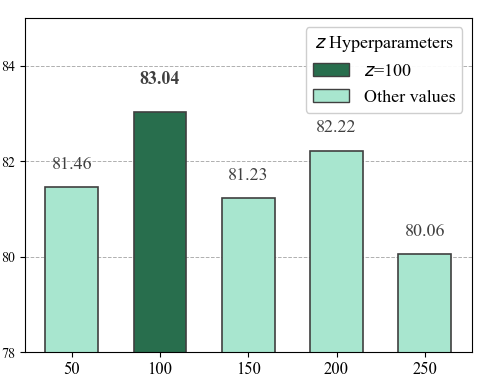}}
\caption{Performance of {\model} using different hyperparameters on SVHN, dir$(\epsilon=0.01)$}
\label{3tu}
\end{figure*}
\begin{table}
\centering
\small
\resizebox{\columnwidth}{!}{%
\begin{tabular}{lccc}  
\toprule
Methods      & SVHN & CIFAR10 & CIFAR100\\ 
\midrule
IFCA          & 81.68 & 53.86 & 30.86 \\
IFCA+IGA     & 82.89   & 55.31 & 32.01   \\ 
\hline
Improvement & $\uparrow 1.21\%$ & $\uparrow 1.45\%$ & $\uparrow 1.15\%$ \\
\hline
\hline
CFL         & 68.16 & 47.60 & 26.33 \\
CFL+IGA     & 69.80   & 49.12 & 28.71   \\ 
\hline
Improvement & $\uparrow 1.64\%$ & $\uparrow 1.52\%$ & $\uparrow 2.38\%$ \\
\hline
\hline
CFL-GP        &69.01 & 48.56 & 26.75 \\
CFL-GP+IGA    & 71.40   &50.62 & 28.80    \\ 
\hline
Improvement & $\uparrow 2.39\%$ & $\uparrow 2.06\%$ & $\uparrow 2.05\%$ \\
\hline
\hline
PACFL       & 80.65 & 55.89 & 30.96 \\
PACFL+IGA    & 84.09   & 58.94 & 34.27 \\ 
\hline
Improvement & $\uparrow 3.44\%$ & $\uparrow 3.05\%$ & $\uparrow 3.31\%$  \\ 
\bottomrule
\end{tabular}%
}
\caption{Compatibility analysis by integrating our proposed IGA into representative CFL methods with $dir(\epsilon=0.01)$.}
\label{or}
\end{table}

\noindent \textbf{Client Participation Scale}. 
We evaluated all approaches across varying client scales to observe how the impact of each algorithm responds to an increasing number of participants. Thus, we set \textit{act}= (0.1, 0.15, 0.2, and 0.25) per training round. \textit{\textbf{(1)}} As shown in Figure \ref{act}, {\model} demonstrates superior performance. \textit{\textbf{(2)}} Increasing $act(\cdot)$ improves the accuracy of all methods, as more local information becomes available per round. \textit{\textbf{(3)}} Compared with FL and DFKD, CFL and {\model} demonstrate a more stable increase in accuracy. This indicates that, under a fixed number of clients, the CFL algorithm should avoid selecting a high client participation rate when adapting to a Dirichlet distribution. 
\vspace{-5pt}
\subsection{Compatibility Study}\label{scala}
\vspace{-3pt}
To validate the compatibility of the proposed aggregation mechanism within CFL frameworks, we integrated it into several existing CFL optimizers and evaluated their accuracy. As shown in Table~\ref{or}, our cross-cluster distillation module consistently enhances the global performance of each CFL method, with the combination involving PACFL achieving the highest test accuracy. Essentially, the clustering algorithm acts as a fundamental component in CFL. Therefore, these results also demonstrate our method's compatibility with various clustering strategies, as it improves their performance while remaining orthogonal to the underlying clustering approach.

\vspace{-5pt}
\subsection{Ablation Study}\label{ada}
\vspace{-3pt}
To assess the necessity of each component, we test accuracy after removing different key modules and loss functions. In all experiments, we use \(Dir(\epsilon=0.01)\) on the SVHN task. For the modules, we remove \textit{GLS}, \textit{GWF}, and IGA, while for the key loss functions, we eliminate $\mathcal{L}_{cf}$ and $\mathcal{L}_{div}$. In Table~\ref{adaw}, after removing the IGA, {\model} reverts to FedAvg, resulting in the most performance degradation. This underscores the critical importance of extracting and migrating cross-cluster information. \textit{GLS} and \textit{GWF} modules cause the model to average samples and integrate identical weights, respectively. Consequently, the global model becomes incapable of addressing differences in data distributions across groups, leading to further performance declines. $\mathcal{L}_{cf}$ and $\mathcal{L}_{div}$ play key roles in pseudo data generation. Dropping $\mathcal{L}_{cf}$ leads to blurred or less realistic outputs, whereas dropping $\mathcal{L}_{div}$ results in less diverse categories.

\begin{table}
    \centering
    \large
    \begin{tabular}{crr}
        \toprule
         & Method                & Accuracy (\%) \\
        \midrule
        Baseline 
            & {\model}               & \textbf{83.04} \\ 
        \midrule
        \multirow{4}{*}{Module}
            & \textit{$-$ GLS}          & 82.23 \\
            & \textit{$-$ GWF}          & 82.14 \\
            & $-$ IGA                  & 72.76 \\
        \midrule
        \multirow{3}{*}{Loss} 
            & $-$$\mathcal{L}_{cf}$   & 81.55 \\
            & $-$$\mathcal{L}_{div}$   & 82.69 \\
        \bottomrule
    \end{tabular}
    \caption{The impact of each module and loss. The experiments are conducted on SVHN, dir$(\epsilon=0.01)$.}
    \label{adaw}
\end{table}

\vspace{-5pt}
\subsection{Hyperparameters Sensitivity Analysis}\label{sensit}
\vspace{-3pt}
We evaluate the sensitivity of key hyperparameters: $\beta_{\text{cf}}$, $\beta_{\text{div}}$, and noise dimension. The $\beta_{\text{cf}}$ and $\beta_{\text{div}}$ weights are assessed at (0.5, 0.75, 1, 1.25, and 1.5), while the noise dimensions are set to (50, 100, 150, 200, and 250), following the settings in~\cite{ftg}. \textit{\textbf{(1)}} As illustrated in Figure~\ref{fig:a} and Figure~\ref{fig:b}, {\model} achieves optimal performance when $\beta_{\text{cf}} = 1$ and $\beta_{\text{div}} = 1$. \textit{\textbf{(2)}} An unsuitable $\beta_{\text{cf}}$ may lead to capturing the group model semantic imbalances in the pseudo-data. Similarly, the improper ratio of $\beta_{\text{div}}$ could reduce the diversity of pseudo-data. \textit{\textbf{(3)}} As shown in Figure~\ref{nz}, a noise dimension of 100 produces the best performance, indicating that excessively high noise dimensions should be avoided in this setup.

\vspace{-8pt}
\section{Conclusion}
\vspace{-3pt}
Training a consensus model that effectively balances personalized and global knowledge under statistical heterogeneity remains a challenge in FL. We address this by introducing a novel framework that dynamically clusters clients to capture personalization and leverages a cluster-level DFKD strategy to distill and transfer global knowledge across groups.
\vspace{-5pt}
\section*{Acknowledgements}
\vspace{-3pt}
Zeqi Leng, Chunxu Zhang and Bo Yang are supported by the National Natural Science Foundation of China underGrant Nos.U22A2098,62172185,62206105, and 62202200; the Major Science and Technology Development Plan of Jilin Province under Grant No. 20240302078GX; the Major Science and Technology Development Plan of Changchun under Grant No.2024WX05. Riting Xia is supported by the Inner Mon-golia University High-level Talent Project under Grant No. 10000-23112101/2861.
\bibliographystyle{named}
\bibliography{FormattingGuidelines-IJCAI-25/FormattingGuidelines-IJCAI-25/arxiv}

\end{document}